# MambaLSTM: A Spatio-Temporal Framework for Enhanced Traffic Accident Risk Prediction

Zhen Yu[1], Yachao Yuan[1,2(✉)], Zixiang Peng[1], Muting Li[1], and Thar Baker[3]

[1] School of Future Science and Engineering, Soochow University, Suzhou, Jiangsu, China
chao910904@suda.edu.cn
[2] Key Laboratory of General Artificial Intelligence and Large Models in Provincial Universities, Soochow University, Suzhou, Jiangsu, China
[3] College of Computing and Intelligent Systems, University of Khorfakkan, Sharjah, UAE

**Abstract.** In traffic accident risk prediction, most studies overlook the extra noise that could be incorporated when fusing temporal features into spatial features, and some models struggle to capture global correlations among spatial regions. To address these challenges, we propose a novel traffic accident risk prediction framework named MambaLSTM. First, we develop a squeeze-and-excitation temporal feature fusion module to integrate temporal information without compromising spatio-temporal integrity. Second, we introduce a new patch embedding module for effectively capturing semantic relationships among spatially adjacent regions. Additionally, we introduce a Mamba block based on state-space models to model global spatial semantics in urban regions. Finally, we propose a MambaLSTM unit to efficiently capture long- and short-term temporal dependencies for identifying dynamic risk patterns. Extensive experiments on real-world datasets demonstrate the proposed model's superiority over state-of-the-art methods. The code is released at https://github.com/Zhenzovo/MambaLSTM.

**Keywords:** Intelligent transportation system, Traffic accident risk prediction, Spatio-temporal modeling.

## 1 Introduction

As traffic volume continues to increase, road traffic safety has become an increasingly prominent issue. Numerous studies have been conducted on the prediction of road traffic accidents based on traditional machine learning [1, 2] and deep learning [3, 4].

Traditional methods typically treat traffic accident counts in a region or a road section as temporal sequences data and predict it using machine learning models like the hidden Markov model. However, these algorithms are prone to failing to effectively capture the complex spatio-temporal correlations between different regions. Deep learning-based approaches [3-5] have been widely used in traffic accident risk prediction. However, these CNN- or LSTM-based methods struggle to adequately model spatial similarity. To address this issue, Graph Convolutional Networks (GCN) [3] and its variant [4] explicitly model a road network as a graph and capture complex spatial dependencies in the road network topology. However, such information is not fully uti-lized for risk prediction in most existing work. C-ViT [6] leverages transformer architecture to capture global spatial semantics for traffic accident risk prediction, but it remains constrained by the balance between predictive accuracy and computational efficiency. Additionally, the occurrence of traffic accidents is often closely tied to specific time slots [7]. Without effective temporal encoding, models typically struggle to directly capture cyclical patterns of particular time periods. Moreover, unsuitable temporal encoding methods introduce additional noise due to interactions between spatial and temporal features [8].

To address the aforementioned issues, we propose *MambaLSTM*, an innovative spatio-temporal architecture for traffic accident risk prediction, which incorporates *Mamba* into *LSTM* architecture to capture global spatial semantics and long-term temporal dependencies efficiently. Specifically, we first design a squeeze-and-excitation for temporal feature fusion module based on channel attention to fuse temporal and spatial features. Second, a patch embedding with multi-scale convolution module is developed to capture adjacent spatial correlations. Then, a MambaLSTM unit is introduced to dynamically model long-term temporal dependencies using a new gating mechanism, while within each MambaLSTM unit, a Mamba block is proposed for capturing global spatial semantic information of road networks.

## 2 Related Work

Traffic risk prediction has consistently remained at the forefront of research, yielding numerous effective methodologies. Early methods include Gaussian kernel-based SVMs [9], regression models [10], and optimized KNN [11] to extract key risk factors. However, these methods are limited by their static assumptions and linear modeling paradigms, making it difficult to capture the highly nonlinear, and dynamic nature of traffic risks, especially under complex spatio-temporal conditions. To overcome this, deep learning models, including CNN, LSTM and GCN [5, 12, 13], have been proposed to extract spatial semantic information for enhancing the prediction model's performance. Despite these advances, existing methods often rely on external graph constructions instead of enhancing the model's intrinsic spatio-temporal learning capabilities.

Spatio-temporal modeling is indispensable for traffic accident risk prediction. Current approaches are mainly based on CNN-RNN hybrid or transformer architectures. CNN-RNN variants, such as ConvLSTM [5] and ST-LSTM [14], improve long short term dependency modeling. Nevertheless, constrained by the local receptive fields of convolutional operations, most of them have difficulties in capturing long-range spatial dependencies. Transformer-based models, such as STTN [15] and Autoformer [16], offer an alternative by enabling global spatio-temporal interactions, but they are often computationally intensive, posing challenges for real-world, large-scale traffic risk prediction applications. Recently, state-space models like Mamba (e.g., GAMMA-Net [17] and UniMamba [18]) have shown superior capability and linear complexity in long-range spatio-temporal modeling. Nevertheless, these methods lack task-specific designs for traffic accident risk prediction. In contrast, we propose a unified MambaLSTM framework to address the unique challenges in accident risk prediction.

## 3 Problem Formulation

**Unit Area.** We divide a city $\mathcal{C}$ into $H \times W$ grid. Consequently, the city is represented as a collection of unit areas $\mathcal{C} = \{c_{1,1}, \dots, c_{i,j}, \dots, c_{H,W}\}$.
**Human Mobility Tensor.** To characterize dynamic urban traffic for accident risk prediction, taxi trajectory data is utilized as a proxy for human mobility. For any given time step $t$, the human mobility across all unit areas is formulated as a tensor $\mathcal{S}^t \in \mathbb{R}^{2 \times \mathbb{H} \times \mathbb{W}}$. Within this tensor, 2 is the flow dimension, which includes the inbound and outbound flow data (i.e., the inflow/outflow of human traffic into/from a unit area).
**Accident Data Tensor.** For any given time step $t$, traffic accident records for all unit areas are organized into a tensor $\mathcal{R}^t \in \mathbb{R}^{2 \times \mathbb{H} \times \mathbb{W}}$, where 2 means that we use two feature dimensions $\mathcal{R}^t_{1,i,j}$ and $\mathcal{R}^t_{2,i,j}$ for risk assessment. Specifically, $\mathcal{R}^t_{1,i,j}$ serves as a binary indicator of accident occurrence within unit area $c_{i,j}$, while $\mathcal{R}^t_{2,i,j}$ quantifies the accident severity in that unit area $c_{i,j}$, as follows:

$$\mathcal{R}^t_{1,i,j} = \begin{cases} 1 & accident\ exists \\ 0 & else \end{cases}, \tag{1}$$

$$\mathcal{R}^t_{2,i,j} = \begin{cases} 1 + 1 \times P_{\text{injured}} + 2 \times P_{\text{dead}} & \mathcal{R}^t_{1,i,j} = 1 \\ 0 & \mathcal{R}^t_{1,i,j} = 0 \end{cases}, \tag{2}$$

where $P_{\text{injured}}$ and $P_{\text{dead}}$ denote the total number of injuries and fatalities, respectively, within unit area $c_{i,j}$ at time step $t$.
**Weather Feature Tensor.** Considering the strong correlation between meteorological conditions and traffic risks, for any given time step $t$, the weather for all unit areas is represented as $\mathcal{E}^t \in \mathbb{R}^{\mathbb{d} \times \mathbb{H} \times \mathbb{W}}$, where $d$ signifies the quantity of weather features.
**Temporal Information.** Crucial temporal attributes are closely linked to traffic anomalies, such as holidays, day of the week, and specific time-of-day periods. For any given time step $t$, these temporal features are represented using one-hot encoding, denoted as $\mathcal{T}^t_{input} \in \mathbb{R}^L$, where $L$ is the dimensionality of the one-hot encoding.
**Problem Definition.** In a city $\mathcal{C}$, given a spatio-temporal feature $\mathcal{X}_{input}$, including human mobility tensor $\mathcal{S}^t$, historical accident data $\mathcal{R}^t$, and weather feature $\mathcal{E}^t$, and temporal information $\mathcal{T}^t_{input}$ for the previous $T$ time steps, our goal is to predict traffic accident risks for the next time step $T+1$. Here, $T = T_k + T_r$, where $T_k$ represents data from the previous time slots and $T_r$ denotes data from the same time slot of previous days/weeks, capturing both short-term and long-term temporal correlations.

## 4 Methodology

The architecture of the proposed MambaLSTM framework is shown in **Fig. 1**. It comprises four core components: a *Squeeze-and-Excitation for Temporal Feature Fusion (SE-TFF)* module, a *Patch Embedding based on Multi-Scale Convolution (PE-MSC)* module, a *Mamba block* module and a *MambaLSTM unit* module. In the following sections, we detail the mechanics of each module.

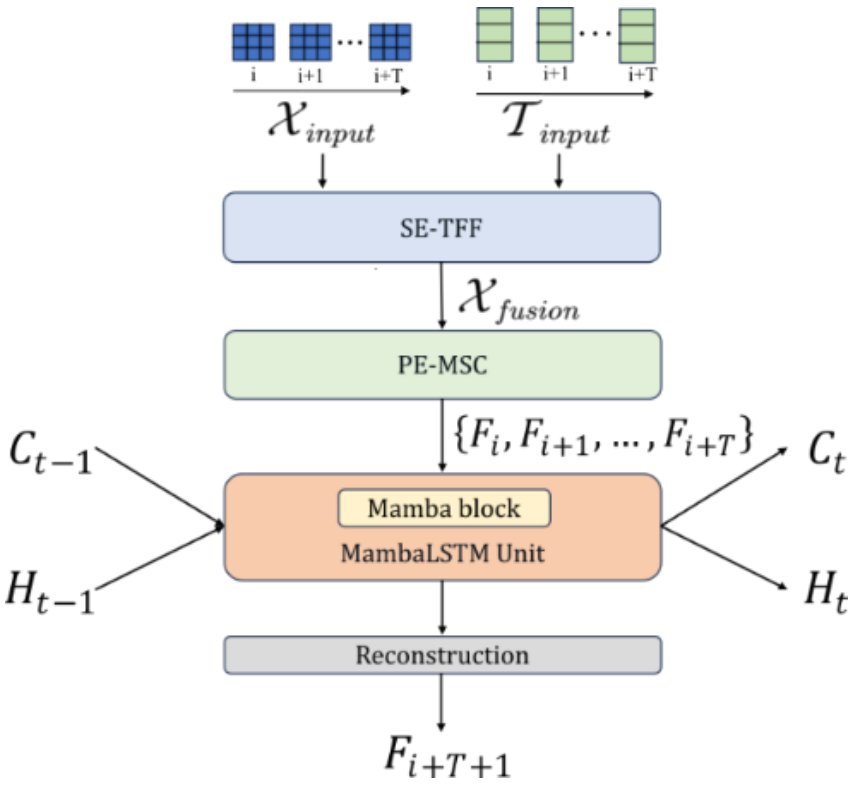

**Fig. 1.** The architecture of the proposed MambaLSTM framework.

### 4.1 Squeeze-and-Excitation for Temporal Feature Fusion (SE-TFF)

During traffic accident risk prediction, improper temporal encoding techniques will introduce extra noise. To address this, we design SE-FEF, a spatio-temporal feature fusion module based on the Squeeze-and-Excitation (SE) attention mechanism [19], enabling attention-driven integration of temporal and spatial features, as shown in **Fig. 2**.

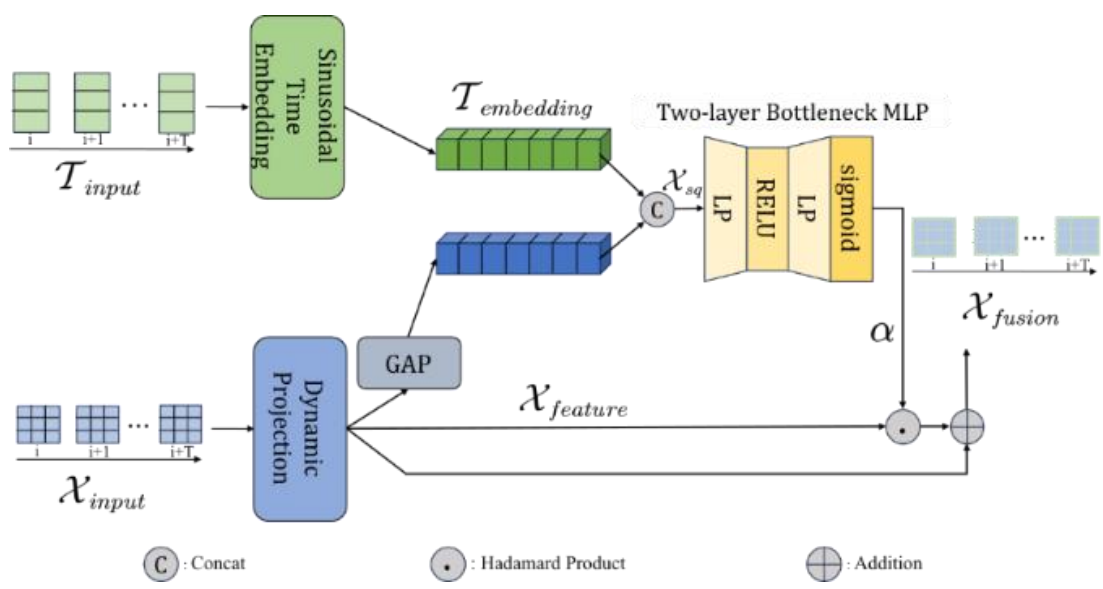

**Fig. 2.** The architecture of the SE-TFF module.

First, the spatio-temporal features $\mathcal{X}_{input}$ are mapped to a high-dimensional latent space through a *Dynamic Projection* layer, generating $\mathcal{X}_{feature}$. Meanwhile, the one-hot encoded temporal features $\mathcal{T}_{input}$ that contain information about hours, days of the week, and holidays, are transformed using *Sinusoidal Time Embedding* to generate a periodic temporal embedding vector $\mathcal{T}_{embedding}$. Then, at each time step $t$, the spatio-temporal feature $\mathcal{X}_{feature}$ is passed through a *Global Average Pooling (GAP)* operation and concatenated with the corresponding temporal embedding vector $\mathcal{T}_{embedding}$. The concatenated result, denoted as $\mathcal{X}_{sq}$, is fed into a *Two-layer Bottleneck MLP* with a channel compression and excitation mechanism, generating the channel attention weights α that encode both spatial and temporal information.

In the *two-layer bottleneck MLP*, the first MLP layer (i.e., LP in **Fig. 2**) reduces the dimensionality of the concatenated feature $\mathcal{X}_{sq}$ and applies a ReLU activation function for nonlinear transformation. The second MLP layer then restores the dimensionality to the original number of channels. Finally, the output is squeezed into a range [0,1] using the Sigmoid function, yielding the scaling factor α.

The scaling factor α is applied to the spatial-temporal feature $\mathcal{X}_{feature}$ in a residual scaling manner, producing the enhanced feature representation $\mathcal{X}_{fusion}$, which incorporates both temporal periodic patterns and spatial contextual semantics. The process can be expressed as follows:

$$\mathcal{X}_{sq} = Concat_{-1}\left(\mathcal{T}_{embedding}, GAP\left(\mathcal{X}_{feature}\right)\right), \tag{3}$$

$$\alpha = \sigma\left(LP_{\left(\frac{C}{r},C\right)}\left(\delta\left(LP_{\left(2C,\frac{C}{r}\right)}\left(\mathcal{X}_{sq}\right)\right)\right)\right), \tag{4}$$

$$\mathcal{X}_{fusion} = (1+\alpha) * \mathcal{X}_{feature}. \tag{5}$$

In this context, $GAP$ stands for global average pooling, while $Concat_{-1}$ represents the merging of two vectors along the last dimension. We define $LP_{(C_{in},C_{out})}$ as a linear mapping that transforms the input size from $C_{in}$ to an output size of $C_{out}$. Additionally, δ and $r$ represent the ReLU activation function and the channel compression ratio, respectively, with $*$ denoting scalar multiplication.

This mechanism selectively integrates periodic patterns while preserving spatial details, enabling robust model performance even during peak traffic periods.

## 4.2 Patch Embedding with Multi-Scale Convolution (PE-MSC)

The prediction of traffic accident risk must account for the semantic correlations of local regions. To handle this, we propose a *Patch Embedding with Multi-Scale Convolution (PE-MSC) module* for $\mathcal{X}_{fusion}$ based on multi-scale hierarchical feature extraction. Details of it are shown in **Fig. 3**.

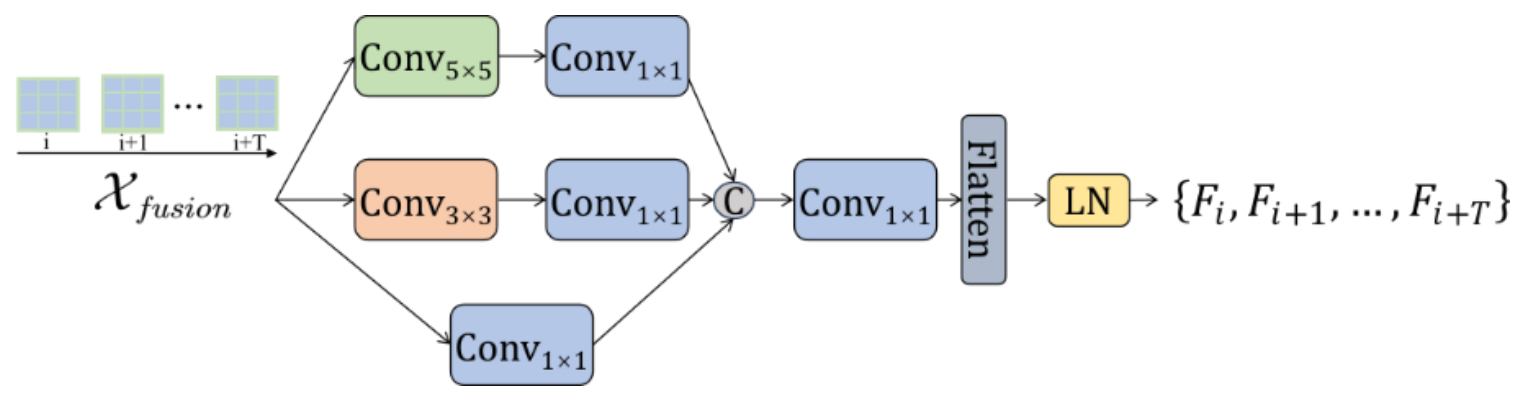

**Fig. 3.** The architecture of the PE-MSC module. LN means Layer Normalization layer.

A hierarchical mechanism for feature extraction is implemented in PE-MSC to model spatial dependencies effectively. Specifically, we employ $3 \times 3$ and $5 \times 5$ convolution kernels in parallel to capture multi-scale spatial features corresponding to different levels of adjacency. The $3 \times 3$ kernel focuses on fine-grained local patterns. In contrast, the 5 × 5 kernel models traffic flow relationships among wider areas, effectively reflecting risk propagation across wider region. Additionally, we introduce a 1 × 1 convolution kernel along each convolution path to perform channel-wise transformations. This operation not only enhances the nonlinear expressive capability but also ensures the integrity of the original feature information through residual connections. After the multi-scale convolution, the outputs are concatenated along the channel dimension (the C symbol in **Fig. 3**), a final 1 × 1 convolution kernel projects the multi-channel features into a unified representation space. Finally, the spatial dimensions are flattened and layer-wise normalized to obtain serialized spatial features $\{F_i, F_{i+1}, \dots, F_{i+T}\}$ across all time steps.

## 4.3 Mamba Block

Despite their outstanding performance for modeling sequential data, SSMs seem less effective at capturing spatial information. This limitation makes them less suitable for traffic accident risk prediction tasks. To handle this, we propose the *Mamba block*, which captures spatial information through a dual-branch parallel processing and feature fusion mechanism, as depicted in **Fig. 4**.

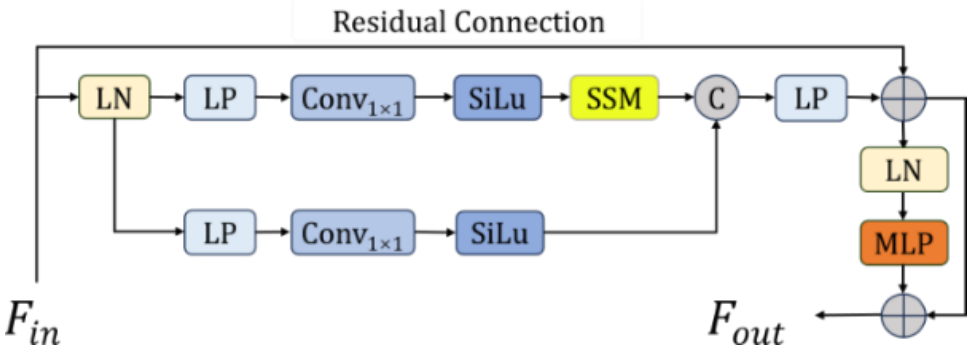

**Fig. 4.** The architecture of the Mamba block. LP stands for Linear Projection.

The Mamba block consists of a two-branch segment, a residual connection, a LP layer, a LN layer, and an MLP. In the two-branch segment, after the initial layer normalization, the first branch applies linear projection and standard 1 × 1 convolution combined with a SiLU activation function, followed by SSMs to model long-range dependencies. To reduce the information loss introduced by the sequential constraints of SSMs, the second branch uses linear projection, standard convolution, and SiLU activation for preserving rich spatial information and enhancing feature expressiveness. The outputs of the two branches are then concatenated along the channel dimension to fuse sequential and spatial features. This fused representation subsequently passes through a linear transformation, another LN layer, and an MLP to reorganize and refine the features, thereby strengthening the model's ability to capture spatial semantics.

Additionally, we introduce a *residual connection* to prevent information loss or degradation due to vanishing gradients during the training of deep networks. Through the Mamba block, we achieve comprehensive parsing of global spatial semantics in traffic accident risk data.

The working process of the Mamba block is formulated as follows:

$$\widetilde{F_{in}} = LN(F_{in}), \tag{6}$$

$$F_1 = SSM\left(SiLu\left(Conv_{1\times1}\left(LP_{\left(C,\frac{C}{2}\right)}\left(\widetilde{F_{in}}\right)\right)\right)\right), \tag{7}$$

$$F_2 = SiLu(Conv_{1\times1}\left(LP_{\left(C,\frac{C}{2}\right)}\left(\widetilde{F_{in}}\right)\right), \tag{8}$$

$$F_{mid} = LP_{(C,C)}\left(Concat_{-1}(F_1, F_2)\right) + F_{in}, \tag{9}$$

$$F_{out} = MLP\left(LN(F_{mid})\right) + F_{mid}. \tag{10}$$

In this context, $F_{in}$ represents the input to the Mamba block, $SSM$ refers to the linear scan operation of SSMs, $LN$ represents layer normalization, $Conv_{1\times1}$ denotes a standard $1 \times 1$ convolution, and $SiLu$ represents the Sigmoid Linear Unit activation function. The Mamba block efficiently models complex spatial correlations within road networks by integrating SSMs into the dual-branch spatio-temporal feature fusion architecture, significantly enhancing the model's understanding of global spatial semantics.

### 4.4 MambaLSTM Unit

We divide a road network into grid units for modeling and analysis. However, due to the complex topology of urban traffic networks, interactions between grid units often span non-adjacent regions. To overcome this, we introduce the Mamba block. The Mamba block leverages a linear-complexity SSM to achieve a global receptive field through efficient hidden state updates, while ensuring computational efficiency. It allows each grid unit to dynamically model its interactions with any region within the road network. Therefore, as shown in **Fig. 5**, we present a novel recurrent unit based on the Mamba block architecture to perform spatio-temporal modeling.

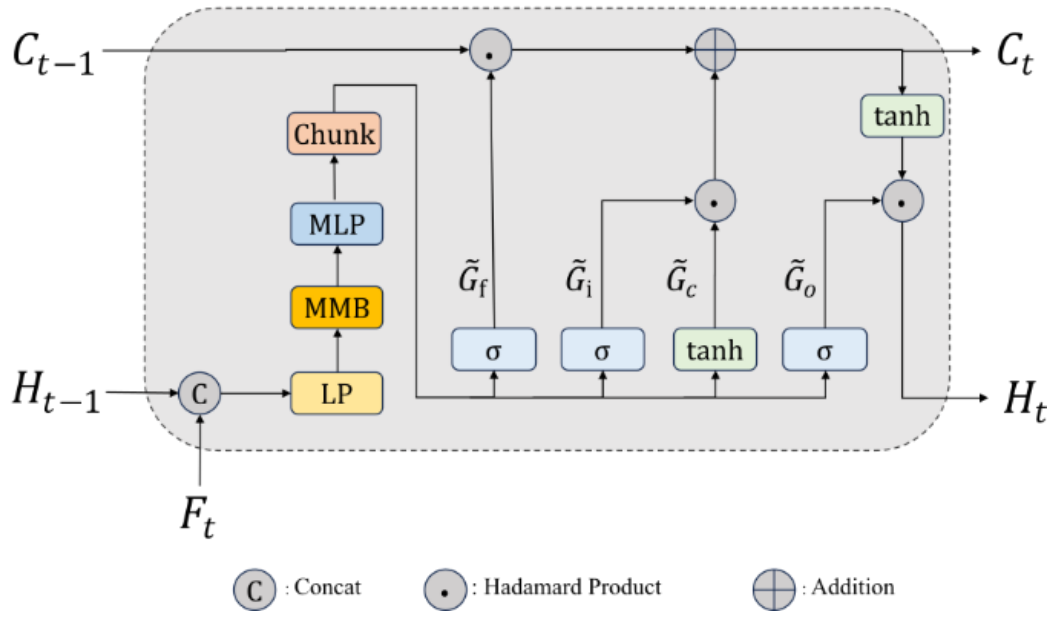

**Fig. 5.** The architecture of the MambaLSTM unit. MMB represents multiple mamba blocks and $\sigma$ denotes a sigmoid activation function.

Building upon ConvLSTM [5], we remove the conventional weight matrices $W$ and biases $b$, and instead incorporate our designed Mamba block. However, directly eliminating the weights and biases leads to structural degradation of the gating mechanism:

due to parameter sharing, the input gate, forget gate, and output gate lose their ability to decouple features, which undermines the model's capability to selectively remember or coordinate multi-dimensional information flows (such as sudden risk signals or spatial heterogeneity) in complex spatio-temporal scenarios like traffic surges and weather changes. To address this, we propose the *MambaLSTM unit*, which captures global spatio-temporal dependencies by reconstructing the LSTM gating mechanism using an MLP, as detailed in the following equations:

$$G = MLP\left(MMB\left(LP_{(C,C)}\left(Concat_{-1}(F_t, H_{t-1})\right)\right)\right), \tag{11}$$

$$\left[G_i, G_f, G_o, G_c\right] = Chunk(G, 4, -1), \tag{12}$$

$$\widetilde{G_x} = \sigma(G_x), x \in (i, f, o), \tag{13}$$

$$\widetilde{G_c} = \tanh(G_c), \tag{14}$$

$$C_t = \widetilde{G_f} \circ C_{t-1} + \widetilde{G_i} \circ \widetilde{G_o}, \tag{15}$$

$$H_t = \widetilde{G_o} \circ \tanh(C_t). \tag{16}$$

Here, $MMB$ represents Multi-Mamba blocks, which are formed by stacking multiple Mamba blocks. $F_t$ represents the fused feature sequence derived from the *PE-MSC* module, while $H_{t-1}$ signifies the hidden state from the prior time step. The operation $Chunk(G, n, -1)$ refers the uniform partitioning of tensor G into $n$ sub-tensors along the last channel dimension. Furthermore, σ denotes the Sigmoid activation function and ∘ represents the Hadamard product.

The output $G$ is divided into four independent sub-tensors $(G_i, G_f, G_o, G_c)$ along the channel dimension using the $Chunk$ operation. These outputted sub-tensors $(G_i, G_f, G_o)$ are then passed through the Sigmoid activation function to generate the input gate $(\widetilde{G_i})$, forget gate $(\widetilde{G_f})$, and output gate $(\widetilde{G_o})$. The candidate cell state $(\widetilde{G_c})$ is calculated by applying the Tanh activation function on the outputted sub-tensor $(G_c)$. The cell state $C_t$ and the hidden state $H_t$ are updated, ensuring that the model adaptively adjusts the retention and forgetting strengths of the information based on the current input features. This improves the model's stability in modeling long-term dependencies.

# 5 Experiment

## 5.1 Experimental Setup

**Datasets.** Our evaluation is conducted across three real-world benchmarks: New York City (NYC), Chicago (CHI) and US-Accidents (USA) datasets. Specifically, NYC encompasses 2013, totaling 147K recorded accidents. CHI contains 44K accident records spanning from February to September 2016. For USA, we use the 500K sampled version and further extract the records from 2022, totaling 113K recorded accidents.

**Implementation Details.** We utilize the Weighted Mean Squared Error (MSE) loss function on traffic accident risk data, which are highly imbalanced. The MSE is defined as:

$$\text{Weighted MSE} = \frac{1}{n}\left[\beta \sum_{y_i=1}(1-\widehat{y_i})^2 + \sum_{y_i=0}\widehat{y_i^2}\right]. \quad (17)$$

Here, β represents the weighting coefficient, and $n$ denotes the number of samples. We set β to 1.25 for NYC and 1.5 for CHI, determined by extensive experiments.

## 5.2 Experimental Results

**Performance Comparison. Table 1** presents the predictive performance of MambaLSTM and baseline models across three evaluation metrics on three real world datasets. MambaLSTM effectively parses complex inter-regional spatial interaction and temporal periodicity by integrating SSM with gating mechanisms, the results demonstrate that MambaLSTM consistently outperforms other models in nearly all metrics.

**Table 1.** Performance Comparison of Different Models.

| Model | NYC (%) | | | CHI (%) | | | USA (%) | | |
|---|---|---|---|---|---|---|---|---|---|
| | F1-score | AUC-PR | AUC-ROC | F1-score | AUC-PR | AUC-ROC | F1-score | AUC-PR | AUC-ROC |
| HA[2] | 37.64 | 51.9 | 57.17 | 16.44 | 20.83 | 54.15 | 13.65 | 10.62 | 50.82 |
| ARIMA[20] | 48.22 | 47.27 | 71.23 | 21.83 | 14.65 | 65.06 | 10.28 | 7.5 | 50 |
| XGBoost[21] | 36.89 | 55.59 | 55.77 | 9.37 | 4.52 | 31.72 | 18.74 | 45.37 | 64.74 |
| LSTM[22] | 44.45 | 57.4 | 83.73 | 13.78 | 33.47 | 84.3 | 41.23 | 48.9 | 86.79 |
| ConvLSTM [5] | 53.37 | 62.14 | 86.8 | 19.81 | 39.12 | 90.9 | 47.07 | 55.7 | 91.65 |
| GSNet[23] | 54.83 | 61.6 | 86.16 | 19.37 | 38.5 | 89.02 | 50.86 | 56.81 | 91.44 |
| ST-GCN[4] | 54.77 | 62.33 | 87.35 | 16.38 | 39.73 | 91.22 | 49.53 | 57.03 | 92.17 |
| MVMT-STN [24] | 54.8 | 62.09 | 86.35 | 16.72 | 39.82 | 90.55 | 50.51 | 57.29 | 91.84 |
| MG-TAR [13] | 53.38 | 61.21 | 86.91 | 18.66 | 39.35 | 91.17 | 40.72 | 44.74 | 89.49 |
| C-ViT[6] | 56.04 | 62.05 | 86.99 | 15.49 | 39.33 | 90.68 | 48.25 | 56.1 | 91.83 |
| ST-TAR[25] | 36.77 | 31.91 | 81.73 | 30.12 | 41.22 | 89.58 | 27.74 | 20.01 | 85.36 |
| Mamba[26] | 56.85 | 60.43 | 86.25 | 31.95 | 34.82 | 88 | 40.11 | 40.14 | 79.95 |
| Ours | 59.52 | 64.09 | 87.83 | 32.92 | 41.27 | 91.58 | 52.2 | 56.77 | 91.72 |

**Visualization Analysis. Fig. 6** visualizes the spatial consistency between MambaLSTM predictions and ground-truth accident occurrences. The predicted maps are smoother than the ground-truth maps and also highlight neighboring areas around accident cells, which is reasonable because traffic accident risk is spatially correlated. These results indicate that MambaLSTM captures meaningful spatial risk patterns.

**Ablation Study.** To ensure that each module of our model is essential for addressing the research problem, we construct three variant models derived from the original MambaLSTM (denoted as M0), i.e., M1, M2, and M3. M1 removes the PE-MSC module, M2 excludes the SE-TFF module, and M3 replaces the MambaLSTM unit module with a ConvLSTM module. In **Fig. 7**, we present the performance of M0, M1, M2, and M3 across three evaluation metrics on the two datasets. The results demonstrate that MambaLSTM outperforms all three variant models across all evaluation metrics, confirming the critical contribution of each component to the model's performance.

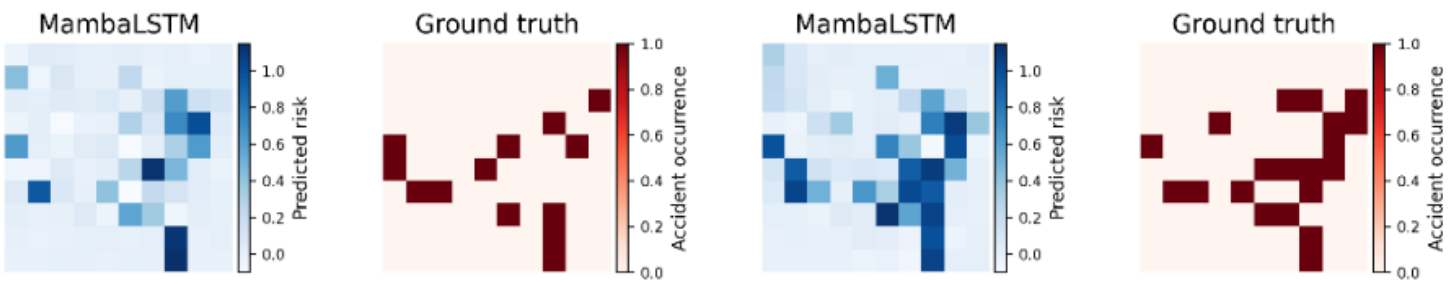

**Fig. 6.** Visualization comparison between MambaLSTM predicted risk heatmaps and ground-truth accident occurrence maps in two cases.

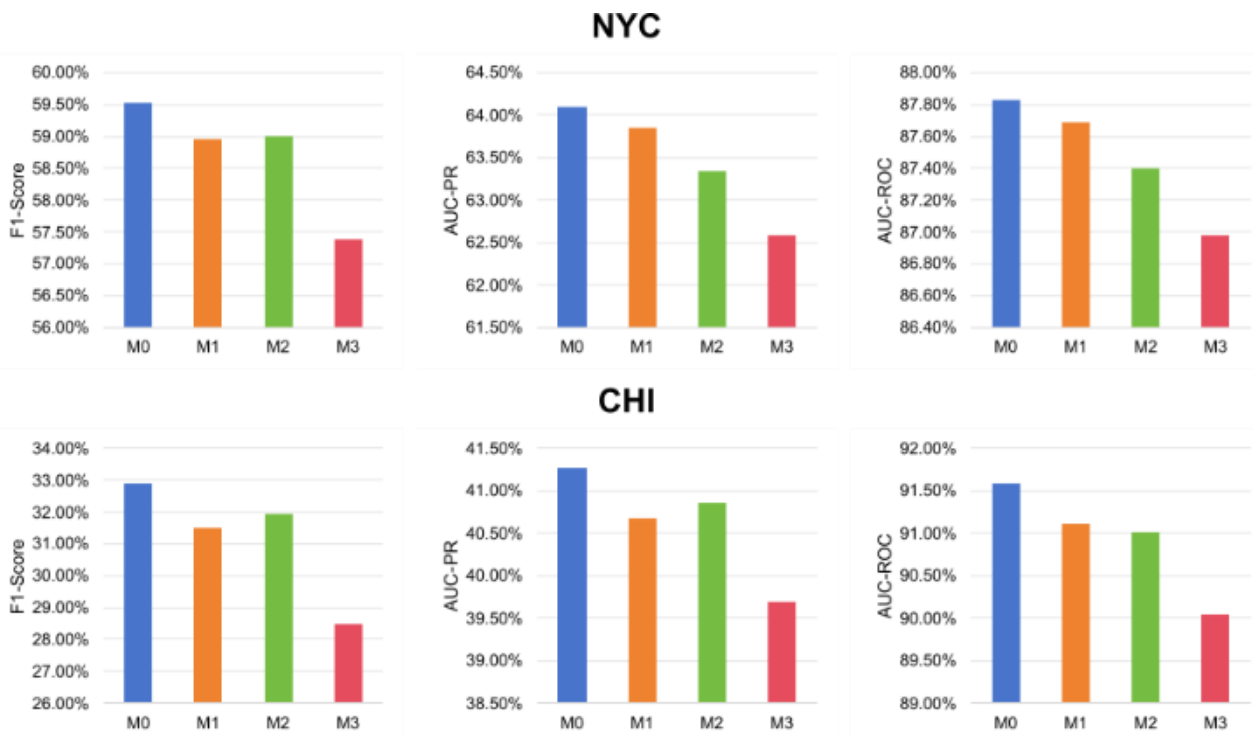

**Fig. 7.** Distribution of three evaluation metrics for the MambaLSTM and its variant models.

**Efficiency Comparison. Table 2** compares the parameter counts and prediction times of MambaLSTM and baseline models on the CHI dataset. Benefiting from the linear computational complexity of SSMs, MambaLSTM requires less training time than MG-TAR and C-ViT. Although its sequential processing of temporal data results in a longer prediction time compared to convolutional models, MambaLSTM completes traffic accident risk prediction for an entire city in 1.76 seconds, which is still sufficiently fast and fully meets the real-time forecasting requirements for urban traffic scenarios.

**Table 2.** Efficiency Comparison: Parameter Counts (/K) and Prediction Times (s) on the CHI Dataset.

| Model | LSTM | ConvLSTM | GSNet | ST-GCN | MVMT-STN | MG-TAR | C-ViT | Ours |
|---|---|---|---|---|---|---|---|---|
| Param | 18.7 | 164.4 | 210.6 | 99.5 | 407.7 | 4349.4 | 670.2 | 402.1 |
| Time | 0.10 | 0.23 | 0.17 | 0.19 | 0.26 | 1.26 | 4.53 | 1.76 |

# 6 Conclusion

In this paper, we proposed an innovative framework named *MambaLSTM*, which can effectively capture both global spatial semantics and long-term temporal dependencies. First, we proposed a squeeze-and-excitation temporal feature fusion module that dynamically integrates temporal signals into spatial representations while mitigating redundant noise. Second, we designed a patch embedding module with multi-scale convolution, which effectively captures local and long-range spatial correlations by modeling semantic relationships among adjacent urban regions. Third, we introduced a Mamba block based on state-space models to encode global spatial semantics, and further integrated it into a newly designed MambaLSTM unit, allowing the model to learn both short- and long-term temporal dependencies through a novel gating mechanism. Extensive experiments on real-world datasets demonstrated the effectiveness, efficiency, and robustness of our approach.